\documentclass[conference]{IEEEtran}
\usepackage[latin9]{inputenc}
\usepackage{amsmath}
\usepackage{graphicx}
\usepackage[numbers,sort&compress]{natbib}
\usepackage{amsfonts}

\ifCLASSINFOpdf
\else
\fi

\usepackage{stfloats}
\begin{document}
%
\title{Scalable Fine-grained Generated Image Classification Based on Deep Metric Learning}

\author{\IEEEauthorblockN{Xinsheng Xuan, Bo Peng, Wei Wang and Jing Dong*\thanks{* Corresponding author.}}
\IEEEauthorblockA{Center for Research on Intelligent Perception and Computing (CRIPAC), 
	\\National Laboratory of Pattern Recognition (NLPR), 
	\\Institute of Automation, Chinese Academy of Sciences (CASIA)\\
Email: xinsheng.xuan@cripac.ia.ac.cn, \{ bo.peng, wwang, jdong \}@nlpr.ia.ac.cn}
}

\IEEEoverridecommandlockouts

%


\maketitle

\begin{figure}[b]
\vspace{-0.3cm}
\parbox{\hsize}{\em
WIFS`2019, December, 9-12, 2019, Delft, Netherlands.
XXX-X-XXXX-XXXX-X/XX/\$XX.00 \ \copyright 2019 IEEE.
}\end{figure}


\begin{abstract}
Recently, generated images could reach very high quality, even human eyes could not tell them apart from real images. Although there are already some methods for detecting generated images in current forensic community, most of these methods are used to detect a single type of generated images. The new types of generated images are emerging one after another, and the existing detection methods cannot cope well. These problems prompted us to propose a scalable framework for multi-class classification based on deep metric learning, which aims to classify the generated images finer. In addition, we have increased the scalability of our framework to cope with the constant emergence of new types of generated images, and through fine-tuning to make the model obtain better detection performance on the new type of generated data. 
\end{abstract}



%
\IEEEpeerreviewmaketitle

\section{Introduction}
\begin{figure}[t]
	\centering
	\includegraphics[width=0.8\linewidth]{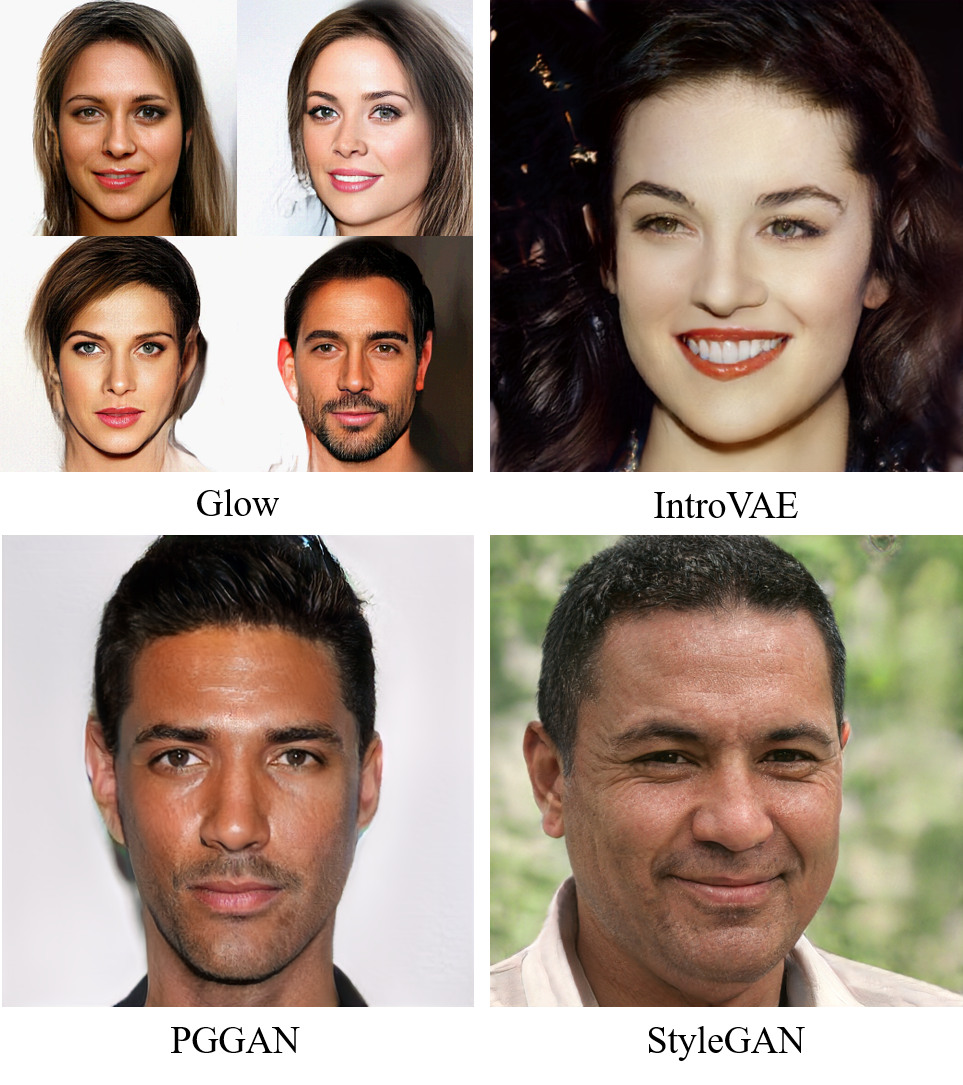}
	\caption{Examples of generated images based on Glow \cite{kingma2018glow}, IntroVAE \cite{huang2018introvae}, PGGAN \cite{karras2017progressive} and StyleGAN \cite{karras2019style}, respectively.}
	\label{gans}
\end{figure}

Over the past few years, the rise of AI-generated images has got a lot of people very worried. Especially since the birth of Generative Adversarial Network (GAN) \cite{goodfellow2014generative}, various GAN-based image generation methods have emerged one after another. According to incomplete statistics in The-GAN-Zoo \footnote{https://github.com/hindupuravinash/the-gan-zoo}, there are currently more than 490 image generation methods for GAN images. In addition to GAN, there are other image generation techniques available, such as variational autoencoders (VAEs) \cite{kingma2013auto} and Glow \cite{kingma2018glow}. Fig.\ref{gans} shows some examples of generated face images, and these high resolution images can be used to fool people. The AP \footnote{https://apnews.com/bc2f19097a4c4fffaa00de6770b8a60d} says it found evidence of a what seems to be a spy using an AI-generated profile picture to fool contacts on LinkedIn, and the fake profile, given the name Katie Jones, connected with a number of policy experts in Washington. So far, advances in image generation technology has had some negative effects. 


Existing detection methods \cite{Mo2018,karras2017progressive,Marra2018,Yang2019,Tariq2018} for generated image is mainly for detecting generated images, and some detection methods treat various types of generated images as a type of fake images. However, these methods ignore that the fingerprint information \cite{marra2019gans,yu2018learning} existing in different types of generated images is different, resulting in the classification result that is not fine-grained. Due to the rapid development of GAN, new types of generated images have emerged soon. Traditional multi-class classification network based on deep learning has fixed the output dimension when training, so the inflexibility of the trained model leads to its poor detection performance for the new types of generated images. Retraining is used as a common practice for dealing with new types of generated images, but this is a time consuming and labor intensive process. 

To solve the above problems, we propose a multi-class classification network architecture based on deep metric learning. Our goal is to classify the fake images finer. We believe that metric learning can force our model learning to map different types of generated image fingerprints \cite{marra2019gans,yu2018learning} in the embedding space. In order to cope with the emergence of new types of generated images, we use the method of feature matching with the Template Image Library (TIL) to achieve scalable and fine-grained classification. Ideally, the feature space learned by our model can be applied to the detection of new types of generated images, but it is difficult to directly use the model to detect new types of images due to the limited number of generated image types. Therefore, we propose to use a smaller number of new types of generated images to fine-tune the trained model, which can further improve the performance of detecting new types of generated images. \textbf{The source code and	data will be available upon acceptance of this paper. }

\begin{figure*}[t]
	\centering
	\includegraphics[width=0.9\linewidth]{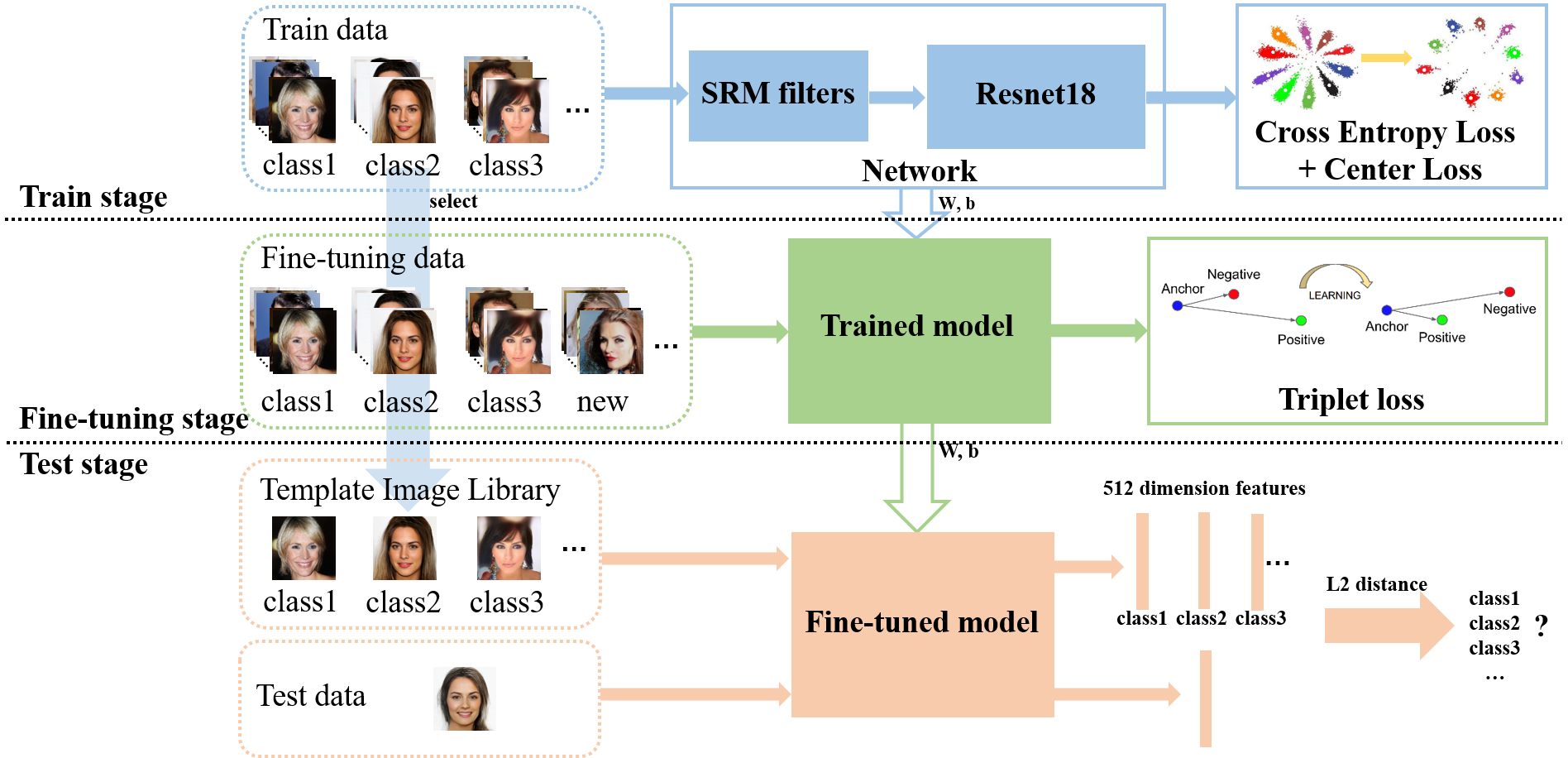}
	\caption{The whole framework consists of three stage: train stage, fine-tuning stage, and test stage. During the training stage, a number of different types of images are sent to the network and trained in conjunction with the center loss based on metric learning. In the fine-tuning stage, we just use new type of few images combined with triplet loss based on metric learning to fine tune the network. For the test stage, we randomly select one of each type of image from the training dataset as Template Image Library (TIL), and send it and test images to our trained model to get the 512-dimensional features. The L2-distance of the test images and the images in the TIL is compared, and the smallest is our model prediction result. }
	\label{overall_framework}
\end{figure*}

\section{Related Work}
The detection of GAN, VAE and Glow images is a new area of image forensics, and there are not many papers in this area. Even so, the researchers have proposed some feasible methods for detecting generated images \cite{marra2019gans,yu2018learning,Li2018,Marra2018,Tariq2018,Mo2018,McCloskey2018,Hsu2019,Li2019,Cozzolino2018}.


Most of the existing detection methods are only for detecting a single type of generated images. For example, Mo et al. \cite{Mo2018} propose a convolutional neural network (CNN) based method to identify fake face images generated by PGGAN \cite{karras2017progressive}. Similarly, Marra et al. \cite{Marra2018} present a study on the detection of images manipulated by GAN-based image-to-image translation, and their results show that deep learning based networks are more robust than traditional methods. Not for a single type of generated images, we take the task of multi-class classification for generated images as our main task. 

In addition to the general methods based on deep learning, there are some methods \cite{cozzolino2018noiseprint,mazumdar2018universal} based on metric learning that are used to enable the network to learn distance information between different data types. Cozzolino and Verdoliva \cite{cozzolino2018noiseprint} propose a method based on deep metric learning to extract a camera model fingerprint, where the scene content is largely suppressed and model-related artifacts are enhanced. In \cite{mazumdar2018universal}, the authors propose a deep learning method based on metric learning which can differentiate between different types of image editing operations. Both \cite{cozzolino2018noiseprint} and \cite{mazumdar2018universal} propose methods based on metric learning, and they focus on tampered images. However, we focus on detection tasks for multiple types of generated image. 

Does GAN have unique feature information similar to human fingerprint information? Francesco et al. \cite{marra2019gans} raised this issue. In their work, they show that each GAN leaves its specific fingerprint in the images it generates, just like real-world cameras mark acquired images with traces of their photo-response non-uniformity pattern. In \cite{yu2018learning}, the authors proposed a deep convolutional neural network for extracting GAN fingerprints and achieved good performance in detecting GAN. Different from the method of Yu et al \cite{yu2018learning}, we propose an scalable architecture based on template matching to cope with the emergence of new types of images. 


\section{Proposed Method}
The overall proposed framework is shown in Fig. \ref{overall_framework}. It is a scalable multi-class classification framework based on deep metric learning. More details are presented in the following. 


\subsection{Network Structure}
Recently, Zhou et al. \cite{Zhou2017,zhou2018learning} used methods based on steganography analysis in their work, using Steganalysis Rich Model (SRM) filters \cite{Fridrich2012} to process images to perform tampering image detection tasks. Inspired by this, we apply SRM filters to the multi-class classification task of generated images. We believe that SRM filters can remove high-level information such as image semantics or image content, and expose the fingerprint information or defect information of generated images, which is advantageous for the classification task. Immediately after SRM filters, we used a residual neural network (ResNet) \cite{he2016deep} to enable our network to directly learn the more effective noise information processed by SRM filters. 


\subsection{Training}
We change the output of the fully connected layer of the last layer to 512 dimensions, which is used in conjunction with center loss based on metric learning, and then adds the fully connected layer for cross entropy loss. Center loss \cite{wen2016discriminative} and cross entropy loss are used in the model training stage, and the combination of the two can make the model learn a feature map with strong discrimination. In fact, the role of center loss here is to minimize the intra-class distance, so that the feature distribution of the same class has strong cohesion. The role of cross entropy loss is to maximize the distance between classes and thus improve the separability between classes. Minimizing the intra-class ariations is critical, here is the center loss function: 

\begin{equation}
\mathcal{L}=\frac{1}{2} \sum_{i=1}^{m}\left\|\boldsymbol{x}_{i}-\boldsymbol{c}_{y_{i}}\right\|_{2}^{2}
\end{equation}

where, \begin{math}\boldsymbol{x}_{i} \in \mathbb{R}^{d}\end{math} denotes the $i$th deep features, belonging to the ${y_{i}}$th class, $d$ is the feature dimension, \begin{math}\boldsymbol{c}_{y_{i}} \in \mathbb{R}^{d}\end{math} denotes the $y_{i}$th class center of deep features. Ideally, the \begin{math}\boldsymbol{c}_{y_{i}}\end{math} should be updated as the deep features changed.

\subsection{Fine-tuning}

In order to further optimize our model's detection performance for new types of generated images, we used an approach based transfer learning to fine tune our trained model. Since the new type of image data is extremely small, we give up using center loss for fine-tuning, but use Triplet loss \cite{Schroff2015}. The reason for using triplet loss is that the number of new types of images used for fine-tuning is small, so we need to make full use of this small number of new types of images. Combining new types of images with other types of images into triples will produce thousands of triples to fine tune our trained model. In addition, triplet loss can be used to minimize intra-class distance and maximize inter-class distance. This is also the case that we have at least two new types of images in our fine-tuning stage, which minimizes the distance between new types of image classes. In short, we can use the triplet loss ensure that an image $x_{i}^{a}$ (anchor) of a specific class is closer to all other images $x_{i}^{p}$ (positive) of the same class than it is to any image $x_{i}^{n}$ (negative) of any other class. Thus we want, 

\begin{equation}
\mathcal{L}=\sum_{i}^{N}\left[\left\|f\left(x_{i}^{a}\right)-f\left(x_{i}^{p}\right)\right\|_{2}^{2}-\left\|f\left(x_{i}^{a}\right)-f\left(x_{i}^{n}\right)\right\|_{2}^{2}+\alpha\right]_{+}
\end{equation}

\begin{equation}
\forall\left(f\left(x_{i}^{a}\right), f\left(x_{i}^{p}\right), f\left(x_{i}^{n}\right)\right) \in \mathcal{T}
\end{equation}

where, $f(x) \in \mathbb{R}^{d}$ is represented as converting an image $x$ to $d$-dimensional features, $\alpha$ is a margin that is enforced between positive and negative pairs,  $\mathcal{T}$ is the set of all possible triplets in the training set and has cardinality $N$. 

\subsection{Testing}
Since our model is based on metric learning, our model predictions depend on the feature distribution of its output. Compared with most other models which directly depend on the model predictions, our method is more scalable. In order to further enhance the practicability of our model, we use a scalable TIL to complete the multi-class classification task. A random sample of each type of image from the training data is placed in TIL, so that the TIL contains the real image and various different types of generated images, and the image type in the TIL can be extended at any time. When testing our multi-class classification model, we compare the features of the tested images with each of the template image in the TIL. Selecting the image type with the smallest L2-distance is the result of the model prediction. Our model has a more explainable side, and it tells you which type of generated image it belongs to, not just the result of real or fake similar to other classification network. 

\begin{table*}[t]
	\begin{center}
		\caption{The performance of our binary classfication network on different datasets}
		\begin{tabular}{|l|c|c|c|c|c|c|c|c|}
			\hline
			Dataset   & BEGAN & DCGAN & Glow  & PGGAN & StyleGAN & IntroVAE & WGANGP & Mixed \\ \hline
			AUROC(\%) & 99.99 & 99.99 & 99.99 & 99.98 & 99.99    & 99.99    & 99.99  & 99.88       \\ \hline
		\end{tabular}
		\label{binary_classification}
	\end{center}
\end{table*}

\section{Experiments}
The CelebFaces Attributes Dataset (CelebA) \cite{liu2015deep} is selected as the dataset of the real image we use, and the image size in the dataset is cropped and resize to 128$\times$128. We collected 7 types of generated image datasets, namely BEGAN \cite{berthelot2017began}, DCGAN \cite{radford2015unsupervised}, Glow \cite{kingma2018glow}, PGGAN \cite{karras2017progressive}, StyleGAN \cite{karras2019style}, IntroVAE \cite{huang2018introvae} and WGANGP \cite{gulrajani2017improved}. Since the original image size of different generated image datasets is different, we uniformly set the image size to 128$\times$128. So our model's input image size is 128$\times$128 with 3 channels. For different types of datasets, we have two ways to get it, one is the way the author provides the dataset to download, such as CelebA and PGGAN. And the other is to use the self-trained model or pre-trained model for image generation, such as DCGAN, WGANGP, Glow, etc. 

\subsection{Binary-classification Experiment}
The purpose of this experiment was to verify that real images and various generated images are separable, providing powerful theoretical support for our subsequent multi-class classification experiments. We believe that the distribution of real images and each type of generated image is separable, so we train binary classifications networks to verify this speculation. Based on the above introduction, we modify the output dimension of our network to 1, and add the sigmoid activation function as the last layer of our network. Optimize cross entropy loss using ADAM optimization algorithm with default parameters. 

CelebA is paired with other images of the generated image type as the training set, and the size of training set of each type of image is set to 10,000, and the size of test set of each type of image is 1,000. In addition, we used 70,000 CelebA images and 70,000 generated images of all types as the training set, and used 10,000 CalebA images and 10,000 images of all types of generated images as test sets. After training based on the above settings, we obtained 7 generated image detection models for a single type and a mixed detection model for all types of generated images. We used the Area Under the Receiver Operating Characteristic curve (AUROC) to measure performance, the test results as shown in TABLE \ref{binary_classification}. For example, the last column of Mixed represents the detection result of the binary classification model trained by CelebA and the mixed generated images. In summary, our model performs well on the detection of a single type of generated images, and it also proves that real images and various generated images are separable. 

As a comparison, we also include performance with different neural network architectures from \cite{Tariq2018} on CelebA and PGGAN datasets. As with the settings in \cite{Tariq2018}, we adjust the image sizes of CelebA and PGGAN to different sizes as input, and finally our test results as shown in TABLE \ref{many_methods}. The results show that our network can achieve better performance compared with previous state-of-the-arts. 

\begin{table}[t]
	\begin{center}
		\caption{AUCROC performance of our method and other methods \cite{Tariq2018} on PGGAN and CelebA}
		\begin{tabular}{|l|c|c|c|c|}
			\hline
			\multicolumn{5}{|c|}{\textbf{AUROC(\%)}}                                                    \\ \hline
			\textbf{Method} & \textbf{64$\times$64} & \textbf{128$\times$128} & \textbf{256$\times$256} & \textbf{1024$\times$1024} \\ \hline
			VGG19           & 56.69          & 55.13            & 57.13            & 60.13              \\ \hline
			XceptionNet     & 79.32          & 79.03            & 82.03            & 85.03              \\ \hline
			NASNet          & 83.55          & 90.55            & 92.55            & 96.55              \\ \hline
			ShallowNetV3    & 90.85          & \textbf{99.99}   & 99.99            & 99.99              \\ \hline
			Our method      & \textbf{99.78} & 99.98            & \textbf{99.99}   & \textbf{99.99}     \\ \hline
		\end{tabular}
		\label{many_methods}
	\end{center}
\end{table}

\begin{table*}[t]
	\begin{center}
		\caption{Close-set test performance of our method and PRNU}
		\begin{tabular}{|l|l|l|l|l|l|l|l|}
			\hline
			Method             & PRNU-1 & PRNU-5 & PRNU-10 & PRNU-20 & PRNU-40 & PRNU-80 & Our method  \\ \hline
			Top-1 Accuracy(\%) & \multicolumn{1}{c|}{55} & \multicolumn{1}{c|}{70} & \multicolumn{1}{c|}{74} & \multicolumn{1}{c|}{81} & \multicolumn{1}{c|}{83} & \multicolumn{1}{c|}{85} & \multicolumn{1}{c|}{\textbf{99}} \\ \hline
		\end{tabular}
		\label{close_set}
	\end{center}
\end{table*}

\subsection{Multi-class Classification Experiment}
For our model training and fine-tuning stages, we complete 10 epochs end-to-end training with an initial learning rate of 0.001 and reduce learning rates after each epoch. ADAM optimization algorithm is used with default parameters, and batch size is set to 128. We set the last layer of fully connected feature output dimensions to be 512. The difference is that we need to optimize the center loss in the training stage and the triplet loss in the fine tuning stage. In addition, we added a fully connected layer which output dimension is the number of training images classes to cater for cross entropy loss function during training stage. In short, during the training stage we used a combination of center loss and cross entropy loss for our total loss function.

We use CelebA images and 7 other types of generated images as our dataset. We classify each type of image into training set and test set. Each class has a training set size of 10,000 images, and each type of test set has a size of 100 images. In the test stage, we compare each test image with each type of template image in TIL, and the one class with the smallest L2-distance is the model prediction result. So we use Top-1 Accuracy to measure performance of our model. Top-1 Accuracy is the conventional accuracy: the model answer (the one with the smallest L2-distance) must be exactly the expected answer.

\subsubsection{Close-set Test}
We chose the PRNU \cite{lukavs2006detecting} method as the baseline comparison method for the test of our multi-class classification model. We used CelebA and other 7 types of generated image datasets as close-set experiments. Similarly, for PRNU-based methods, our datasets were used for testing of the PRNU method. The result is as shown in TABLE \ref{close_set}. For example, the column of PRNU-5 represents the use of 1,000 images as test set and using 5 images to extract a fingerprint template, and the column of our method represents the use of 1,000 images as test set and 1 images as template image. The results show that our method can achieve good performance in detection of many types of generated images, and is superior to traditional methods.

\begin{table*}[t]
	\begin{center}	
		\caption{Performance of our method in open-set test and different numbers of fine-tuned image tests}
		\begin{tabular}{|l|c|c|c|c|c|c|c|c|c|c|c|c|c|c|}
			\hline
			\multicolumn{15}{|c|}{\textbf{Top-1 Accuracy(\%)}} \\ \hline
			Dataset & \multicolumn{2}{c|}{BEGAN} & \multicolumn{2}{c|}{DCGAN} & \multicolumn{2}{c|}{Glow} & \multicolumn{2}{c|}{PGGAN} & \multicolumn{2}{c|}{StyleGAN} & \multicolumn{2}{c|}{IntroVAE} & \multicolumn{2}{c|}{WGANGP} \\ \hline
			Before fine-tuning & 99 & 99 & 99 & 58 & 99 & 45 & 99 & 66 & 93 & 53 & 99 & 31 & 99 & 90 \\ \hline
			5 images to fine-tuning & 99 & 99 & 99 & 98 & 99 & 99 & 99 & 16 & 99 & 10 & 99 & 3 & 99 & 91 \\ \hline
			10 images to fine-tuning & 99 & 99 & 99 & 99 & 99 & 99 & 99 & 75 & 99 & 61 & 99 & 15 & 99 & 99 \\ \hline
			20 images to fine-tuning & 99 & 99 & 99 & 99 & 99 & 99 & 99 & 82 & 99 & 96 & 99 & 84 & 99 & 99 \\ \hline
			40 images to fine-tuning & 99 & 99 & 99 & 99 & 99 & 99 & 99 & 88 & 99 & 94 & 99 & 92 & 99 & 99 \\ \hline
			80 images to fine-tuning & 99 & 99 & 99 & 99 & 99 & 99 & 99 & 91 & 99 & 99 & 99 & 98 & 99 & 99 \\ \hline
		\end{tabular}
		\label{open_set}
	\end{center}
\end{table*}

\begin{table*}[t]
	\begin{center}	
	\caption{Test performance for the scalability of our method when progressively adding new types of generated images}
	\begin{tabular}{|l|l|l|l|l|l|l|l|l|l|l|l|l|l|}
		\hline
		Dataset & BEGAN+CelebA & \multicolumn{2}{c|}{Add DCGAN} & \multicolumn{2}{c|}{Add Glow} & \multicolumn{2}{c|}{Add PGGAN} & \multicolumn{2}{c|}{Add StyleGAN} & \multicolumn{2}{c|}{Add IntroVAE} & \multicolumn{2}{c|}{Add WGANGP} \\ \hline
		Top-1 Accuracy(\%) & 99 & 99 & 99 & 99 & 99 & 98 & 92 & 92 & 97 & 85 & 86 & 84 & 97 \\ \hline
	\end{tabular}
	\label{scable_test}
	\end{center}
\end{table*}

\subsubsection{Open-set and Fine-tuning Test}
In response to the emergence of new types of generated images, we designed experiments for open-set testing and fine-tuning testing. The scalability of our method enhances the model's ability to detect new types of generated images. For the model training stage, we combined the idea of k-fold cross-validation. Leave-one-out cross-validation (LOOCV) involves using one observations as the test set and the remaining observations as the training set. Similarly, we keep a type of generated image marked as $A$ not trained, using the remaining type of generated image for training, but the final test set contains $A$. In this way, we trained the model in combination with CelebA images and six of seven types of generated images, and the remaining type of generated image is used as the test set. 

\begin{figure}[t]
	\centering
	\includegraphics[width=0.9\linewidth]{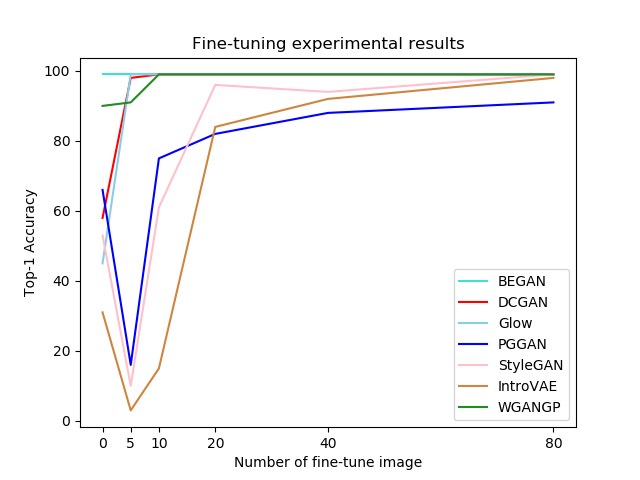}
	\caption{Results after different numbers of generated image fine-tuning}
	\label{fine_tune}
\end{figure}

In order to further enhance the model's detection performance for new types of generated images, we fine-tuned our model and tested it. The results of Top-1 Accuracy are shown in the TABLE \ref{open_set}. The first row of data represents the results of the open-set test before fine-tuning. Since the performance of the open-set test can be improved, we fine-tuned our trained model. Each column represents a different number of new types of generated images we use to fine tune our trained model. For example, the two columns in where BEGAN is located indicates that the model did not use the BEGAN data set during training, with the left column representing the average detection performance of all test data except BEGAN, and the right column representing the detection performance of only the BEGAN data set. It can be seen that the detection tasks of high quality images (such as PGGAN, StyleGAN, IntroVAE) require more fine-tuning images to achieve good performance compared to low quality images (such as BEGAN, DCGAN). 

For a more intuitive observation, the corresponding results of the table are shown in the Fig.\ref{fine_tune}. The x-axis indicates that different numbers of images are used as few-shot tests, and the y-axis indicates the detection results of the new type of generated images by the fine-tuned model. When the number of generated images used to fine tune the network is 5, the fine-tuning performance of the high quality image is degraded. We believe that the reason is that high quality generated images are more deceptive, so that more image data is needed to fine-tune the model to force the model to fit a reasonable image distribution. When the number of fine-tuned images is 20, the detection performance of the fine-tuned model performs well. 

\subsubsection{Scability Test}

We set up experiments to verify the scalability of our method, and the test performance is shown in the TABLE \ref{scable_test}. 10,000 BEGAN images and 10,000 CelebA images are used to train the model, and its performance is shown in the BEGAN+CelebA column, and then the trained model is fine-tuned by increasing the type of generated image one by one, and its performance is shown in other columns. For example, the DCGAN column indicates that 20 DCGAN images are used to fine tune our trained model, and the Glow column indicates that 20 DCGAN images and 20 Glow images are used to fine tune our trained model. The DCGAN column contains two sub-columns, the left side representing the average detection performance of all test data except DCGAN, and the right side representing the detection performance of only the DCGAN data set. As the number of types of generated images increases, the probability of misclassification of our models increases, but overall our method performs well in scalability. 

\section{Conclusion}
In this paper, we focus on the multi-class classification problem of multiple types of generated images. Based on different types of generated images, there are different fingerprint information. We propose a scalable framework for multi-class classification based on deep metric learning. After verifying that real images and the various types of generated images are separable, we tested our multi-class classification model performance. In order to verify the performance of the model's scalability, we used new type of generated images to test our model, and finally improved its performance through fine-tuning experiments. 





\bibliographystyle{IEEEtran}
\bibliography{IEEEexample}
%

%
%
%
%
%

\end{document}